%% file: main.tex
\PassOptionsToPackage{table,xcdraw}{xcolor}
\documentclass[sigconf,authorversion,nonacm]{acmart}

\AtBeginDocument{%
  }

\setcopyright{acmlicensed}
\copyrightyear{2024}
\acmYear{2024}
\acmDOI{XXXXXXX.XXXXXXX}

\acmConference[CIKM '24]{33rd ACM International Conference on
Information and Knowledge Management}{October 21--25,
  2018}{Boise, Idaho, USA}
\author{Federico Belotti}
\affiliation{%
  \institution{DISCo, University of Milan-Bicocca}
  \city{Milan}
  \country{Italy}}
\email{federico.belotti@unimib.it}
\orcid{0009-0008-0140-3318}

\author{Fabio Dadda}
\affiliation{%
  \institution{DISCo, University of Milan-Bicocca}
  \city{Milan}
  \country{Italy}}
\email{fabio.dadda@unimib.it}

\author{Marco Cremaschi}
\affiliation{%
  \institution{DISCo, University of Milan-Bicocca}
  \city{Milan}
  \country{Italy}}
\email{marco.cremaschi@unimib.it}

\author{Roberto Avogadro}
\affiliation{%
  \institution{SINTEF}
  \city{Oslo}
  \country{Norway}}
\email{roberto.avogadro@sintef.no}

\author{Matteo Palmonari}
\affiliation{%
  \institution{DISCo, University of Milan-Bicocca}
  \city{Milan}
  \country{Italy}}
\email{matteo.palmonari@unimib.it}

\usepackage{graphicx}
\usepackage[nolist]{acronym}
\usepackage{xspace}
\usepackage{hyperref}
\usepackage{tabularx}
\usepackage{multicol}
\usepackage{multirow}
\usepackage{adjustbox}
\usepackage{arydshln}
\usepackage{numprint}
\usepackage{todonotes}
\usepackage{listings}
\usepackage{amsmath}
\usepackage{float}
\usepackage{tikz}
\usepackage{caption}
\usepackage[compact]{titlesec}
\usetikzlibrary{tikzmark, calc}
\usepackage[normalem]{ulem}
\usepackage[english]{babel}
\usepackage{microtype}
\usepackage{tikz}
\usepackage{colortbl}
\usepackage{makecell}
\usetikzlibrary{backgrounds}
\usepackage{hyperref}
\usepackage{tablefootnote}

\useunder{\uline}{\ul}{}

\newcommand{\ie}{\textit{i}.\textit{e}.,\xspace}
\newcommand{\eg}{\textit{e}.\textit{g}.,\xspace}

\newcommand{\lamapi}{\textit{LamAPI}\xspace}
\newcommand{\selbat}{\textsf{s-elBat}\xspace}
\newcommand{\turl}{\textsf{TURL}\xspace}
\newcommand{\tablellama}{\textsf{TableLlama}\xspace}
\newcommand{\llamatwo}{\textsf{Llama2}\xspace}
\newcommand{\alligator}{\textsf{Alligator}\xspace}
\newcommand{\dagobah}{\textsf{Dagobah}\xspace}

\definecolor{in}{HTML}{caffbf}
\definecolor{mood}{HTML}{fdffb6}
\definecolor{ood}{HTML}{ffd6a5}

\begin{document}

\title{Evaluating LLMs on Entity Disambiguation in Tables}

\begin{abstract}
Tables are crucial containers of information, but understanding their meaning may be challenging. 
% Indeed, recently, there has been a focus on Semantic Table Interpretation (STI), \ie the task that involves the semantic annotation of tabular data to disambiguate their meaning. 
Over the years, there has been a surge in interest in data-driven approaches based on deep learning that have increasingly been combined with heuristic-based ones. In the last period, the advent of \acf{llms} has led to a new category of approaches for table annotation.
% The interest in this research field has led to many approaches employing different techniques. 
However, these approaches have not been consistently evaluated on a common ground, making evaluation and comparison difficult.
This work proposes an extensive evaluation of four STI \ac{sota} approaches — \alligator (formerly \selbat), \dagobah, \turl, and \tablellama; the first two belong to the family of heuristic-based algorithms, while the others are respectively encoder-only and decoder-only \ac{llms}. We also include in the evaluation both GPT-4o and GPT-4o-mini, since they excel in various public benchmarks. The primary objective is to measure the ability of these approaches to solve the entity disambiguation task with respect to both the performance achieved on a common-ground evaluation setting and the computational and cost requirements involved, with the ultimate aim of charting new research paths in the field.
\end{abstract}




% \received{17 Maggio 2024}
% \received[revised]{}
% \received[accepted]{}

\maketitle

\section{Introduction}\label{sec:introduction}

Tables are commonly used to create, organise, and share information in various knowledge-intensive processes and applications in business and science. Disambiguating values occurring in the table cells using a background \ac{kg} is useful in different applications. First, it is part of the broader objective of letting machines understand the table content, which has been addressed by different research communities with slightly different formulations like \acf{sti}~\cite{Liu2023} - the one considered in this paper, semantic labelling~\cite{pham2016semantic}, and table annotation~\cite{Suhara2022}. The main idea behind these efforts is to match the table against a background knowledge graph by annotating cells (mentions) with entities (Cell-Entity Annotation - CEA), columns with class labels (Column-Type Annotation - CTA), and pairs of columns with properties (Column-Property Annotation - CPA)~\cite{Jimenez2020}. Second, annotations produced by table-to-graph matching algorithms can be used to transform the tables into \ac{kgs} or populate existing ones. Third, links from cells to entities of \ac{kgs} support data enrichment processes by serving as bridges to augment the table content with additional information~\cite{Cremaschi2020-1}. 
This conceptualisation covers most of the proposed definitions by generalizing some aspects (\eg~consideration of NILs and selection of column pairs to annotate).

In particular, we focus on \ac{el} in tables (CEA, in the \ac{sti} terminology), which is relevant not only to support table understanding but also to support data transformation, integration and enrichment processes, which are particularly interesting from a data management point of view. The \ac{cea} task can be broken down into two sub-tasks: candidate \ac{er}, where a set of candidates for each mention is collected and, often, associated with an initial score and rank, and \ac{ed}, where the best candidate is selected (and, in some case, a decision whether to link or not is also considered~\cite{Avogadro2023}). When considering approaches to \ac{sti} and, especially, \ac{cea}, it should be considered that the content and the structure of tables may differ significantly, also depending on application-specific features: column headers may have interpretable labels or be omitted; the number of rows can vary from a dozen (\eg as typical in tables published on the web or in scientific papers) to hundreds thousand or even millions (\eg in business data); the cells may include reference to well-known entities (\eg geographical entities) as well as to specific ones (\eg biological taxa); tables may come with a rich textual context (\eg caption or other descriptions in web or scientific documents), or no context at all (\eg in business data)~\cite{Liu2023}.

A first generation of approaches for \ac{cea} has exploited different matching heuristics, traditional machine learning approaches based on engineered features (in the following ``feature-based ML''), or a combination of both~\cite{Liu2023}.     
\begin{figure*}[t!]
  \centering
  \caption{Architectures of \tablellama, \turl and \alligator.}
  \Description{Architectures of \tablellama, \turl and \alligator.}
  \includegraphics[width=0.8\textwidth]{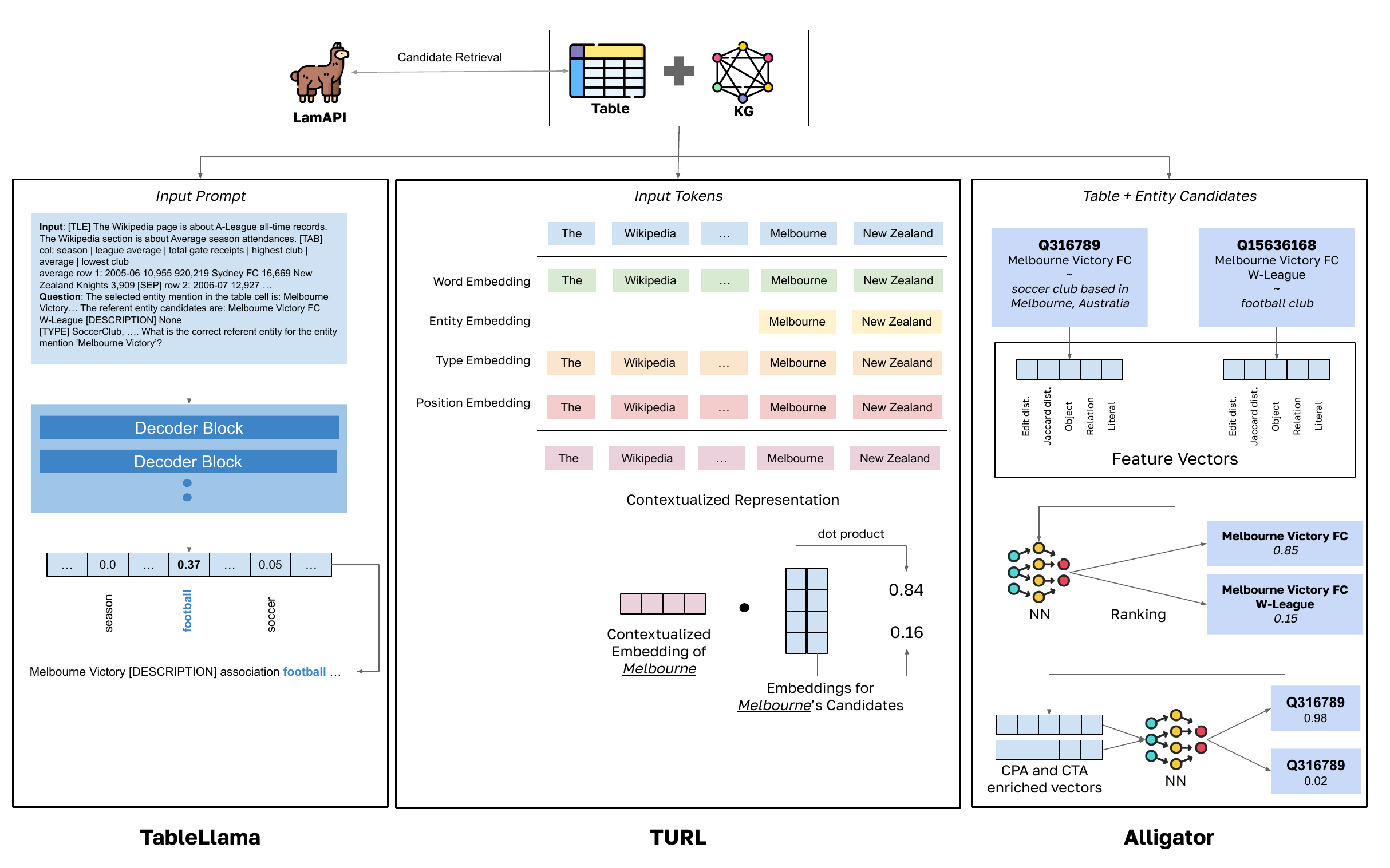}
  \label{fig:architectures}
\end{figure*}
The increased recognition of the power of \ac{llms} has led to a new generation of \ac{gtum} approaches that support \ac{sti} (CEA, CTA, and CPA) and other tasks (\eg question answering, schema augmentation, row population, and fact verification among the others). The first remarkable example is \turl~\cite{Deng2022}, which is based on an adaptation of BERT~\cite{Devlin2019} to consider the tabular structure, additionally fine-tuned to execute specific tasks (including \ac{ed}). The latest of these models, \tablellama~\cite{Zhang2023}, is based on the autoregressive \ac{llm} Llama 2-7B~\cite{Touvron2023-2}, which is fine-tuned with instruction tuning\cite{wei2022instruction} to perform specific tasks. These approaches have been trained and tested on datasets that fit their generalistic ambition and reported as \ac{sota} approaches for the considered tasks. 

While both STI-related and GTUM approaches facilitate entity disambiguation (ED), a comparative evaluation of these two method families on a common experimental foundation has yet to be conducted. STI-related methods have primarily been assessed within the International Semantic Web Challenge on Tabular Data to Knowledge Graph Matching (SemTab), where most tables are synthetically generated by querying knowledge graphs (KGs), closely resembling tables from industrial or corporate environments. These tables, however, lack the rich semantic context of web tables found on platforms like Wikipedia. In contrast, GTUM approaches have been evaluated mainly on web-based tables, which are semantically rich and crafted by humans for human use. Consequently, it remains unclear how these approaches perform relative to the state of the art (SOTA) under different conditions.
Finally, several aspects of \ac{cea} that are important from a data management point of view, \eg scalability, have not been considered. We posit that several of these questions are important to understand the current status of \ac{ed} algorithms for tables, as well as the challenges to be addressed in the future.

With our study, we want to provide a more detailed analysis of the advantages and disadvantages of GTUM \ac{ed} approaches based on \ac{llms}, especially when compared to those specifically focusing on CEA developed in the context of \ac{sti}. Our analysis mainly considers the \ac{sota} generalistic generative model \tablellama~\cite{Zhang2023}, its predecessor \turl~\cite{Deng2022}, a BERT-based encoder-only transformer, and \alligator~\cite{Avogadro2023}, a recent feature-based \ac{ml} approach. These models represent three different inference mechanisms for \ac{ed}, sketched in Figure~\ref{fig:architectures}, each one associated with a set of expected advantages: \tablellama is expected to exploit implicit knowledge of a large \ac{llm}; \turl is expected to be a more efficient generalistic model based on a small \ac{llm} fine-tuned specifically for the \ac{ed} task; \alligator exploits a set of features engineered based on the experience with the SemTab challenge, which is further processed by two small neural networks to return confidence scores. Each approach is tested in \textit{in-domain}, \textit{out-of-domain}, and \textit{moderately out-of-domain} settings, as more precisely defined in Section~\ref{sec:distribution-aware}. In addition, we test the moderately out-of-domain fine-tuning to test generalisation and adaptation capabilities of \turl, \tablellama and \alligator. Finally, we provide the results obtained by \dagobah, an approach that achieved top performance in previous SemTab challenges\footnote{With some limitations due to its excessive execution time and the difficulty of properly replicating the published results despite our best effort.}.

We remark that this paper aims to provide insights on the behavior and impact of GTUM models on \ac{ed} rather than introducing a new approach. This is inspired by the large body of similar analyses addressing specific problems in NLP~\cite{Wang2023,kirk2024understanding,zhong2024can,zhang2024llmeval,kandpal2023llmstruggle}. In our work, we also develop an evaluation protocol that can be used in future research.

To summarise, our \textbf{main contributions} are: 

\begin{enumerate}
    \item Test the performance of different genres of STI models when used in combination with a realistic candidate retrieval method;
    \item Test the performance of these models on several datasets used to evaluate \ac{sota} \ac{sti} approaches, through a common-ground comparison with approaches trained on these data;
    \item Assess the performance improvement by adaptation with additional moderately-out-of-domain fine-tuning (Sec. \ref{sec:distribution-aware});
    \item Evaluate the computational efficiency and provide implications for actual usage in different application settings.
\end{enumerate}

This paper is organised as follows:
Section~\ref{sec:sota} proposes a detailed examination of the techniques used by \ac{sti} approaches in the \ac{sota}. Section \ref{sec:tested-approaches} introduces and details the approaches tested in this work, relating them to the \ac{ed} challenges they are intended to solve. Section~\ref{sec:setup} describes the objectives of this study, the datasets used to evaluate the selected approaches and defines the experimental settings followed in Section~\ref{sec:evaluation}, which introduces the configuration parameters and the evaluation results, discussing the main results and the ablation studies. Finally, we conclude this paper and discuss the future direction in Section~\ref{sec:conclusions}.

\section{Related Work}
\label{sec:sota}
\ac{cea} is usually divided into two sub-tasks: \textit{retrieval of candidate entities}, \ie \ac{er}~\cite{Ratinov2009}, and \textit{entity disambiguation}, \ie \ac{ed}~\cite{Rao2013}, where one or no candidate is selected as a link, usually after scoring and ranking the candidates. Approaches to \ac{ed} for tables have been mostly proposed as part of broader \ac{sti} systems~\cite{Liu2023}. For this paper, we group these approaches into two main categories: i) those based on heuristics, (feature-based) \ac{ml} and probabilistic approaches, and ii) those based on \ac{llm}. 

\textbf{Heuristic, ML and Probabilistic-based Approaches}. We refer to~\cite{Liu2023} for a thorough review of \ac{sti} approaches and \ac{ed} approaches proposed therein up to 2022. The \ac{ed} task in \ac{sti} can be performed by applying multiple techniques while focusing on different inherent information: i) \textit{similarity}, ii) \textit{contextual information}, iii) \textit{ML techniques}, and iv) \textit{probabilistic models}. Often, the disambiguation step involves selecting the winning candidate based on \textbf{heuristics}, like the string \textit{similarity} between the entity label and mention~\cite{Limaye2010,Zhang2017,Oliveira2019,Kruit2019,Steenwinckel2019,Thawani2019,Cremaschi2019,Cremaschi2020-1,Cremaschi2020-2,Huynh2020,Tyagi2020,Shigapov2020,Avogadro2021,Abdelmageed2021,Huynh2021,Chen2022,Cremaschi2022}. 

\textit{Contextual information} during the \ac{cea} task considers the surrounding context of a table cell, such as neighbouring cells, column headers, or header row. Contextual information provides additional clues or hints about the meaning and intent of the mention. By analysing the context, a system can better understand the semantics of the cell and make more accurate annotations~\cite{Syed2010,Ritze2015,Efthymiou2017,Chen2019,Nguyen2019,Thawani2019,Nguyen2020,Nguyen2021,Morikawa2019,Azzi2020,Chen2020,Cremaschi2020-2,Abdelmageed2020,Huynh2020,Avogadro2021,Baazouzi2021,Huynh2021,Huynh2022,Chen2022,Cremaschi2022,Liu2022}.

Other methods that can be employed are \textbf{\ac{ml} techniques}. These techniques typically involve training a \ac{ml} model on a labelled dataset where cells are annotated with their corresponding entities. The models, like Support Vector Machine (SVM)~\cite{Mulwad2010}, Neural Network (NN)~\cite{Thawani2019} and Random Forest~\cite{Zhang2020}, learn patterns and relationships between the cells content and their associated entities. To predict the most appropriate entity, \ac{ml} techniques consider various cell features, such as the textual content, context, neighbouring cells, and other relevant information. \alligator is a recent approach belonging to this family~\cite{Avogadro2023}; it is used in this study and further described in Section~\ref{sec:tested-approaches}.  

\textbf{Probabilistic models} are frameworks for representing and reasoning under uncertainty using probability theory. These models vary in their representation of dependencies and use diverse graphical structures. Several \ac{pgm} can also be used to resolve the disambiguation task, such as Markov models or Loopy Belief Propagation (LBP)~\cite{Mulwad2011,Mulwad2013, Bhagavatula2015,Kruit2019,Yang2021}.

\textbf{LLMs-based approaches.} In the current \ac{sota}, several attempts to apply \ac{llm} in the \ac{sti} process can be identified. Based on the architecture structure of \ac{llms}, approaches can be categorised into two groups: i) \textit{encoder-based}, and ii) \textit{decoder-based}~\cite{Pan2024}. 

Starting from \textit{encoder-based approaches}, Ditto~\cite{Li2020} utilizes Tran\-sformer-based language models to perform a slightly different task, \ie entity-matching between different tables. Doduo~\cite{Suhara2022} performs \ac{cta} and CPA using a pre-trained language model, precisely fine-tuning a BERT model on serialised tabular data. Column (columns-properties) types are predicted using a dense layer followed by an output layer with a size corresponding to the number of column (columns-properties) types. \dagobah SL 2022~\cite{Huynh2022} employs an ELECTRA-based~\cite{electra2020} cross-encoder, a variant of the BERT model, which takes a concatenated input, including table headers and the entity description, and outputs a probability value representing the entity's likelihood concerning the headers. TorchicTab~\cite{Dasoulas2023} is composed of two sub-systems:  TorchicTab-Heuristic and TorchicTab-Classification, with the latter that utilises Doduo~\cite{Suhara2022} internally. 

Regarding the \textit{decoder-only approaches}, \cite{Korini2023} explored the \ac{cta} task by employing ChatGPT, performing experiments with diverse prompts tailored for the task using a subset of the SOTAB benchmark~\cite{Korini2022}. Another study evaluates GPT3.5-turbo-0301 in zero-shot settings on a task that was somehow related to \ac{cea}; the task consisted of classifying descriptions of products based on attribute-value pairs~\cite{Peeters2023}. Some works included structured tabular data into the training process of general purpose decoder-based \ac{llms}  to address the peculiarities of specific domains, \eg BloombergGPT~\cite{Wu2023} for the financial domain or CancerGPT for the medical one~\cite{li2024cancergpt}. Other works, instead, focused on table-specific tasks, especially reasoning~\cite{suiWSDM2024, kong2024opentab, nahid2024tabsqlifyenhancingreasoningcapabilities}, enhanced by Chain-of-Thought (CoT)~\cite{wang2024chainoftable, zheng-etal-2023-chain}. Another decoder-based approach, TableGPT~\cite{Li2023}, performs several tasks, including CTA using a fine-tuned version of GPT-3.5.

Of particular interest for our work, LLM-based models capable of addressing all the STI tasks and in particular the CEA one are: i) \turl~\cite{Deng2022}, which leverages a pre-trained TinyBERT~\cite{Jiao2020} model to initialise a structure-aware Transformer encoder, fine-tuned to obtain contextualised representations for each cell, with matching scores between the \ac{kg} candidates' representations and cell embeddings that are calculated using a linear function, and transformed into a probability distribution over the candidate entities; ii) TableLlama~\cite{Zhang2023}, a fine-tuned \llamatwo on a multi-task dataset for tabular data, performs \ac{cea} ( along with several other tasks), where the entity linking sub-dataset derives from the \turl~\cite{Deng2022} dataset, by prompting the LLM to retrieve the correct candidate given the serialized table, the table metadata, the cell to be linked and a set of proposed candidates.

A more comprehensive review of both \textit{encoder-only} and \textit{decoder-only} LLM-based approaches for Table Understanding can be found in~\cite{fang2024llm-table-survey}.

\input{tables/pre_training_statistics}

\section{Selected Approaches}
\label{sec:tested-approaches}

We select four STI approaches representative of different categories of algorithms that solve semantic tasks on tables, and, especially CEA: \tablellama is the first autoregressive \ac{llm} that is specifically instruction-tuned on tabular data and reports \ac{sota} results~\cite{Zhang2023}; \turl, cited as previous \ac{sota} in~\cite{Zhang2023}, is a BERT-based encoder-only model performing the three STI tasks, including CEA~\cite{Deng2022}; \alligator is a recent feature-based \ac{ml} algorithm focusing on CEA, is publicly available, and has been evaluated in settings similar to the moderately-out-of-domain settings discussed in this paper~\cite{Avogadro2023}; \dagobah, a heuristic-based algorithm, has been the winner of various rounds of the SemTab challenge in 2020~\cite{Huynh2020}, 2021~\cite{Huynh2021} and 2022~\cite{Huynh2022} and is publicly available.

While we focus our study only on the CEA task, we remark that all the selected approaches perform all three STI tasks (which is rare and not obvious, especially for the LLM-based approaches), \ie CTA, CPA and CEA.

\textbf{\turl}~\cite{Deng2022} introduces the standard BERT pre-training/fine\-tuning paradigm for relational Web tables and is composed of three main modules: i) an embedding layer to convert different components of a table into embeddings, ii) a pre-trained TinyBERT Transformer~\cite{Jiao2020} with structure-awareness to capture both the textual information and relational knowledge, and iii) a final projection layer for pre-training/fine-tuning objectives. The embedding layer is responsible for embedding the entire table, distinguishing between word embeddings  $\mathbf{x}_w$, derived from table metadata and cell text (mentions), and entity embeddings $\mathbf{x}_e$, representing the unique entities to be linked. The sequence of tokens in $\mathbf{x}_w$ and the entity representation $\mathbf{x}_e$ are sent to a structure-aware TinyBERT Transformer to obtain a contextualised representation. To inject the table structural information into the contextualised representations, a so-called visibility-matrix $M$ is created, such that entities and text content in the same row or column are visible to each other, except table metadata that are visible to all table components. During pre-training, both the standard \textit{Masked Language Model} (MLM) and the newly introduced \textit{Masked Entity Retrieval} (MER) objectives are employed, with the projection layer that is learned to retrieve both the masked tokens and entities\footnote{The entities are retrieved from a small candidates set, considering that entity vocabulary could be quite large}. During fine-tuning, the model is specialised to address the specific downstream task. Specifically, for CEA, when provided with the sub-table containing all mentions to be linked to an external KG, \turl is fine-tuned to produce a probability distribution over the combined set of candidates for each mention to be linked. This means that \turl lacks the context provided by all the not-to-be-linked cells, without the possibility of abstaining from answering or responding with a NIL. Also, in the original paper, we observe that the best candidate is chosen by a function that compares the best score computed by \turl and the score assigned by the model to the first candidate retrieved by the Wikidata-Lookup service by down-weighting the best model prediction. As in previous comparisons~\cite{Zhang2023}, to evaluate \turl's performance on the \ac{ed} task, we consider the model predictions only. Finally, as \turl accepts the whole table as input, computational performance varies depending on the size of the input table.

\textbf{\tablellama}~\cite{Zhang2023} employs \llamatwo~\cite{Touvron2023-2} and instruction tuning~\cite{wei2022instruction} to solve multiple tasks related to tables, \eg CTA, CPA, CEA and Q\&A to name a few. To this end a multi-task instruction tuning dataset named \textsf{TableInstruct} is made available, containing more than 1.24M training tables and 2.6M instances gathered from 14 different table datasets of 11 distinctive tasks. To account for longer tables, exceeding the maximum context length of \llamatwo (which is fixed at 4096 tokens), LongLora~\cite{Chen2024} is used to enlarge the maximum context length to 8192 tokens. In particular, the LLM is prompted with i) an instruction that describes the high-level task to be solved, ii) an input prepended by the \textsf{[TLE]} special character followed by the table metadata, if available, and the serialised table and iii) a question based on the task to be solved by the LLM. In particular, the table starts with the \textsf{[TAB]} special character and is followed first by the table header and then by every row in the table, separated by the special character separator \textsf{[SEP]}. For the CEA task, the question asks the LLM to link a particular mention found in the table against a small set of candidates (maximum 50 candidates) by extracting the correct candidate from the proposed list without, as in \turl, the possibility of not answer or to answer with a NIL. Since the context is fixed to 8192 tokens, compromises must be made to reduce either the length of the set of candidates or the input table. 
% An example prompt for the CEA task is presented in Table~\ref{fig:tablellama-prompt}.

\begin{table*}[ht!]
    \centering
    \label{tab:approach-comparison}
    \caption{Comparison of \turl, \tablellama, and \alligator for the CEA task.}
        \resizebox{0.9\linewidth}{!}{%
        \begin{tabular}{p{2cm}p{5cm}p{5cm}p{5cm}}
            \toprule
             & \textbf{\turl~\cite{Deng2022}} & \textbf{\tablellama~\cite{Zhang2023}} & \textbf{\alligator~\cite{Avogadro2023}} \\
            \midrule
            \textbf{Model} & TinyBERT~\cite{Jiao2020} equipped with table structure-awareness (visibility matrix) & \llamatwo with LongLora~\cite{Chen2024} for extended context & Feature-based ML approach with two \ac{nn}s for scoring \\
            \midrule
            \textbf{Input} & Table comprised of only the mentions to be linked, optional metadata and candidates for every mentions & Serialized table with instruction, optional metadata, mention to be linked and the list of candidates, limited to 8192 tokens & Whole table (candidates are retrieved on-the-fly thanks to \lamapi) \\
            \midrule
            % \textbf{Special Tokens} & Visibility matrix to capture relational structure & [TLE] for table metadata, [TAB] for table start, [SEP] for rows & No special tokens, features derived from table structure \\
            % \midrule
            \textbf{Objective} & Pre-training with MLM and MER; task-specific fine-tuning on CEA & Multi-task instruction tuning for CEA, CTA, CPA, and other table-related tasks & Feature-based binary cross-entropy with negatives sampling: ${p(y_m=1|F_m) \geq p(y_m=0|F_w;w\in\mathcal{N}(m))}$\footnotemark \\
            \midrule
            \textbf{Output} & Probability distribution over candidates & (Auto-regressively) Extracts candidate directly from list & Score in $[0,1]$ for every candidate \\
            \midrule
            \textbf{Limitations} & Input table limited to mentions to be linked, no NIL-awareness & Limited to 8192 tokens, no score over candidates, no NIL-awareness & No semantic-related features, biased towards \ac{kg} matching\\
            \bottomrule
        \end{tabular}
        }
\end{table*}
\footnotetext{Given a mention $m$ and its correct associated entity in the knowledge graph, $p(y_m = 1 \mid F_m)$ represents the probability that the features $ F_m $ correctly identify the correct candidate. For each incorrect candidate $ w $ in the set $ \mathcal{N}(m) $, $ p(y_m = 0 \mid F_w) $ represents the probability that the features $ F_w $ correctly indicate that $ w $ is not the correct candidate.}

\textbf{\alligator} is a feature-based \ac{ml} approach based on multiple steps: i) \textit{data analysis and pre-processing}, ii) \textit{ER}, iii) \textit{local feature extraction}, iv) \textit{local scoring}, v) \textit{feature enrichment}, vi) \textit{context-based scoring}. 
%and vii) \textit{cell-level confidence estimation}.
The \textit{data analysis and pre-processing} step applies standard pre-processing transformations (\eg lowercase, space removing), with
%converts all cells to lowercase and removes extra spaces and special characters, e.g., underscores (\_), to improve the results of the entity retrieval phase. In addition, 
columns classified as either \textit{L-column} (containing literals) or \textit{NE-column} (containing named-entity mentions).
The \ac{er} step extracts relevant candidates from the \ac{kg} using the \lamapi services. The \textit{local feature extraction} step builds a vector of engineered features for each candidate entity; the vector represents information about the specific candidate (e.g., its \ac{kg} popularity, number of tokens and the syntactical similarity between the candidate label and the mention) or its comparison with the row cells content (\eg trying to match related literals with the row content). \textit{Local scoring} computes a matching score for each candidate, using a simple deep \ac{nn} (denoted PN in~\cite{Avogadro2023}): the \ac{nn} takes the local feature vectors as input and is trained to convert the features into a normalised matching score. In practice, this step re-ranks all the candidates, considering mainly text similarity and matches against values on the same \textit{row}. The last steps compute updated scores for the candidate of a given mention by considering the best candidates for other cells on the same \textit{column}. In practice, \ac{cpa} and \ac{cta} tasks are executed on a best-effort basis to capture, for each candidate, the degree of agreement between its types and properties and the types of properties of other best candidates in the same column. After this \textit{feature enrichment} step, a simple deep \ac{nn} (denoted RN in~\cite{Avogadro2023}) similar to the previous one predicts the updated normalised scores for all the candidates\footnote{In principle, the approach also estimates a confidence score for each cell to make a decision whether to link the best candidate or not; however, in this paper, we focus on the \ac{ed} task and consider the candidate with the best score as the output of \alligator}.

\textbf{\dagobah}~\cite{Huynh2021} performs the \ac{cea}, \ac{cpa} and \ac{cta} tasks by implementing a multi-step pipeline: i) table pre-processing, ii) \ac{er}, iii) candidate pre-scoring, iv) \ac{cpa}, \ac{cta}, and v) \ac{cea}. The \textit{table pre-processing} step involves estimating table metadata, \eg header detection, and column type estimation. These tasks help identify the structure and annotation targets, facilitating subsequent annotation steps. \ac{er} retrieves relevant candidate entities from the \ac{kg} for each mention in the table using Elasticsearch; it considers the primitive types identified during pre-processing and enriches entity information with aliases to increase coverage. Then, the \textit{candidate pre-scoring} computes a relevance score for each candidate by combining two features: context and literal similarity. Given a mention to disambiguate and a candidate entity, the context feature compares the mention neighbors to the candidate neighbors with respect to the underline \ac{kg}. 
%Overall, this method improves the precision of context scoring by incorporating both direct and indirect connections while avoiding overly complex and noisy paths. Then, \ac{cpa} and \ac{cta} modules identify the most suitable relations and types for column pairs and target columns using a majority voting strategy. 
Finally, the most relevant entity for each mention is selected by combining pre-scoring with information from \ac{cpa} and \ac{cta} into a final score. The main challenges with \dagobah are its high time complexity and memory usage, making it impractical for processing large tables. This difficulty stems from the expensive task of matching table rows with relations in a \ac{kg}. Consequently, we limited our evaluation of \dagobah to the datasets \textit{HT-R1}, \textit{HT-R2}, and \textit{WikidataTables-R1} because these datasets contain tables with limited size, as detailed in Section~\ref{sec:datasets}.

Table~\ref{tab:data-pretrain} lists the datasets used to pre-train the selected approaches; in particular, for \tablellama we use the published model on HugginFace, while we train \turl and \alligator on the same datasets used in the original paper. We use the term "pre-train" to refer to this \textit{main} training phase to distinguish it from subsequent fine-tuning (cfr. Section~\ref{sec:training}). \dagobah is not listed because is not based on training.

\input{tables/datasets_statistics}

\section{Study Set-up}
\label{sec:setup}

The main objective of this work is to compare GTUM and feature-based ML approaches on the CEA task within a well-defined and common evaluation setting. 
As specified by the SemTab challenge \cite{hassanzadeh2023results}, the metrics adopted to evaluate an approach are: $\text{Precision (P)} = \frac{\#\; correct \; annotations}{\#\; submitted \; annotations}$, $\text{Recall (R)} = \frac{\#\; correct \; annotations}{\#\; ground-truth \; annotations}$ and $\text{F1} = \frac{2PR}{P + R}$. Since we are mainly interested in measuring the ability of the models to disambiguate the correct entity among a limited set of relevant candidates, \textbf{we ensure the correct entity is always present in the candidates set, and otherwise, we inject it}, as in previous work~\cite{Deng2022,Zhang2023}. Under this setting, Precision, Recall and F1 are the same and equal to the Accuracy metric. Therefore, we report the accuracy as $\text{Acc} = \frac{\#\; correct \; annotations} {\#\; mentions \; to \; annotate}$.

To maintain a good balance between speed, coverage, and diversity, \textbf{we retrieve and inject at most $50$ candidates for every mention}, eventually adding the correct one.

Our evaluation has the main objective of evaluating the selected approaches on datasets not considered in the original experiments, considering especially the evaluation of \turl and \tablellama on STI-derived datasets and of \alligator and \dagobah (with some limitations) on the datasets used to evaluate the first two approaches. Therefore, we first discuss the datasets used in our study (Section~\ref{sec:datasets}), and the evaluation protocol with its associated objectives (Section~\ref{sec:datasets}). Then, we provide details about ER (Section~\ref{sec:er}) and the training of the models (Section~\ref{sec:training}).

\subsection{Datasets}
\label{sec:datasets}

The datasets considered in this work come from different sources and contain information about different domains. In particular, we have selected the following:

\begin{itemize}
    \item \textbf{SemTab2021 - R3 (BioDiv)}~\cite{abdelmageed2021biodivtab}: the BioDiv dataset is a domain-specific benchmark comprising $50$ tables from bio\-diversity research extracted from original tabular data sources; annotations have been manually curated.
    \item \textbf{SemTab2022 - R2 (2T)}~\cite{Cutrona2020}: the dataset consists of a set of high-quality manually-curated tables with non-obviously linkable cells, \ie where mentions are ambiguous names, typos, and misspelled entity names;/
    \item \textbf{SemTab2022 - R1 \& R2 (HardTables (HT))}~\cite{abdelmageed2022results}: datasets with tables generated using SPARQL queries~\cite{Jimenez2020}. The datasets used from HardTables 2022 are \textit{round 1} (R1) and \textit{round 2} (R2);
    \item \textbf{SemTab2023 - R1 (WikidataTables)}~\cite{hassanzadeh2023results}: datasets with tables generated using SPARQL queries for creating realistic\-looking tables. The dataset includes \textit{Test} and \textit{Validation} tables, yet we exclusively employ the \textit{Validation} tables due to \ac{gs} not being provided;
    \item \textbf{TURL}~\cite{Deng2022}: the authors of \turl developed the \turl dataset using the extensive WikiTable corpus, a rich compilation of tables from Wikipedia. This dataset includes table metadata such as the table name, caption, and column headers. This dataset has been sub-sampled by \tablellama's authors to create a smaller version containing exactly 2K mentions, called \textit{TURL-2K}, and used as the test set.
\end{itemize}
Table~\ref{tab:data-stats} reports the statistics of each dataset used for testing and fine-tuning the selected approaches (\textdagger indicates the datasets that have undergone a reduction). The SemTab datasets were already split in train and test except for \textit{BioDiv 2021}, which has been only used during the testing phase along with the \textit{TURL-2K}. To create a unified and common ex\-pe\-ri\-men\-tal setting, each dataset has been modified to contain the same set of mentions: datasets for \tablellama were first created by generating prompts for each mention while filtering out all prompts that exceeded \tablellama's context length (which is equal to $8192$ tokens). For this reason, we have renamed \textit{BioDiv}, \textit{2T} and \textit{TURL-2K} into \textit{BioDiv-red}, \textit{2T-red} and \textit{TURL-2K-red}. To reduce the number of tokens generated for each prompt, we have taken some precautions: i) the \textit{description} separator has been reduced from \textit{[DESCRIPTION]} to \textit{[DESC]}, and ii) the row separator (\textit{[SEP]}) has been deleted. Then, \turl datasets and the ground truths for \alligator and \dagobah were created using the same mentions as \tablellama.

\input{tables/datasets_domain}

\subsection{Distribution-aware Experimental Objectives}
\label{sec:distribution-aware}

The datasets used for training or evaluating \ac{ed} are generated from different sources, contain tables of different sizes, and hold information about disparate domains. %and are generated from different sources. 
%In practice, w
We assume each dataset is associated with a data distribution generating it~\cite{Jimenez2019, Jimenez2020,Cutrona2020,abdelmageed2021biodivtab,hassanzadeh2023results, Deng2022}.

We introduce the concepts of ``\textit{in-domain}'', ``\textit{out-of-domain}'' and ``\textit{moderately-out-of-domain}'' settings related to the evaluation of a particular approach on a test set, specialising a distinction between ``\textit{in-domain}'' and ``\textit{out-of-domain}'' used in~\cite{Zhang2023}. These denominations depend on the source the data has been generated from and the domain(s) it holds the information about. In particular, we define:

\begin{itemize}
    \item ``\textit{in-domain}'' (IN): a test set $Y$ for an approach $A$ is considered ``\textit{in-domain}'' for $A$ if $A$ has been trained on a dataset $X$ genera\-ted from the \textit{same} data source and covering \textit{similar} domain(s) as $Y$;
    \item ``\textit{out-of-domain}'' (OOD): a test set $Y$ for an approach $A$ is considered ``\textit{out-of-domain}'' if $A$ has been trained on a set of data $X$ generated from a \textit{different} data source and covering \textit{different} domain(s) as $Y$;
    \item ``\textit{moderately-out-of-domain}'' (MOOD): a test set $Y$ for an ap\-proach $A$ is considered ``\textit{moderately-out-of-domain}'' for $A$ if $A$ has been trained on a set of data $X$ generated from the \textit{same} data source but covering \textit{different} domain(s) as $Y$ or vice versa, i.e., if $A$ has been trained on a set of data $X$ generated from a \textit{different} source but covering \textit{similar} domain(s) as $X$. This covers a setting such as the evaluation of \tablellama, pre-trained on the \textbf{TURL} dataset~\cite{Deng2022}, on the STI-derived test set \textbf{2T}: the two datasets contain different tables but cover cross-domain information linked to Wikidata in a quite similar way.  
\end{itemize}

Given the definitions above, the evaluation procedure has been divided into two steps

\begin{enumerate}
    \item We assess the performance of the consi\-dered approaches on the test data defined in Section~\ref{sec:datasets} and detailed in Table~\ref{tab:data-stats}, without further fine-tuning. The heterogeneity of the test data implies all the algo\-rithms are tested against ``\textit{in-domain}", ``\textit{moderately-\-out-of-domain}" and ``\textit{out-of-domain}" data.
    \item We fine-tune \turl, \tablellama and \alligator on the train splits from their MOOD data, \ie on SemTab2022 HT-R1, SemTab2022 HT-R2, SemTab2023 WikidataTables-R1 and SemTab2022 2T-red for \turl and \tablellama, and on TURL\-120k\footnote{TURL-120k has been created starting from the original TURL dataset~\cite{Deng2022} by sub-sampling 13'061 tables containing 120'000 mentions ($\approx$ 125'000 is the number of mentions used to fine-tune \tablellama and \turl on their MOOD data combined).} for \alligator; finally we test the fine-tuned models on our test data.
\end{enumerate}

The \textbf{main objectives} of our analysis can be summarised as follows:
\begin{itemize}
    \item test the capability of pre-trained approaches on tables from different datasets, which we expect to be repre\-sentative of different data distributions;
    \item test the generalisation capability of both feature-based ML and LLM-based approaches after fine-tuning in MOOD settings;
    \item compare the approaches in inference time and occupied memory to get insights about their appli\-cability on applica\-tion domains where processing of large tables may be re\-quired (e.g., enrichment of business data).
    \item identify strengths and weaknesses of \ac{llm}-based approaches on the \ac{ed} task with  ablation studies.
\end{itemize}

\subsection{Candidate entity retrieval with LamAPI}\label{sec:er}

The candidates for the mentions contained in the TURL dataset~\cite{Deng2022}, along with its sub-sampled version TURL-2K~\cite{Zhang2023}, were retrieved through the Wikidata-Lookup-Service\footnote{\url{https://www.wikidata.org/w/api.php?action=wbsearchentities&search=Obama&language=en&limit=50&format=json}, which retrieves at most $50$ candidates for the mention ``Obama"}, which is known to have a low coverage w.r.t. other ERs~\cite{Avogadro2022}. For this reason, we researched to identify a state-of-the-art approach/tool specific to the \ac{er}. The final choice fell on \lamapi, an \ac{er} system developed to query and filter entities in a \ac{kg} by applying complex string-matching algorithms.
% As suggested in the paper~\cite{Avogadro2022}, we have integrated DBpedia (v. 2016-10 and v. 2022.03.01) and Wikidata (v. 20220708), which are the most popular \ac{kg}s also adopted in the SemTab challenge\footnote{\href{https://www.cs.ox.ac.uk/isg/challenges/sem-tab}{www.cs.ox.ac.uk/isg/challenges/sem-tab}}.
In \lamapi, an ElasticSearch\footnote{\href{https://www.elastic.co/}{www.elastic.co}} index of the Wikidata dump (v. 01072024) has been constructed, leveraging an engine designed to search and analyse extensive data volumes in nearly real-time swiftly. 
%These customised local copies of the \ac{kgs} are then used to create endpoints to provide \ac{er} services. The advantage is that these services can work on partitions of the original \ac{kgs} to improve performance by saving time and using fewer resources. This simulates an application setting of large-scale entity disambiguation (large tables), where a local copy can speed up operations substantially. The \lamapi \textit{Lookup} service was used to extract the candidates, as carried out by other services~\cite{Avogadro2023}. Given a string input, the service retrieves a set of candidate entities from the reference \ac{kg}.
% \begin{figure}[h!]
%   \centering
%   \caption{Number of mentions w.r.t. number of candidates for the TURL - 2K dataset.}
%   \Description{Number of mentions w.r.t. number of candidates for the TURL - 2K dataset.}
%   \includegraphics[width=\columnwidth]{imgs/turl-wikidata-vs-lamapi.png}
%   \label{fig:lamapi-vs-wikidata}
% \end{figure}
The choice to use \lamapi as an \ac{er} system is based on its availability and performance compared to other available systems (\eg Wikidata Lookup)~\cite{Avogadro2022}. To further validate the choice of \lamapi we have computed the number of mentions with $K$ candidates for the TURL-2K-red dataset: the original TURL-2K dataset has almost $600$ mentions with $1$ candidate (the correct one, with $969$ mentions ($\approx 48\%$) with at most 5 candidates included the correct one). For all those mentions the Wikidata-Lookup service fails to retrieve something meaningful. On the contrary \lamapi retrieves for $1650$ mentions ($\approx 91\%$) in our sub-sampled dataset at least $45$ candidates. In particular, for the TURL-2K-red dataset, the coverage (\ie how many times the correct candidate is retrieved by the \ac{er} system over the total number of mentions to cover) of \lamapi and Wikidata-Lookup is $88.17\%$ and $71.75\%$ respectively. For all these reasons we decided to replace the candidates extracted from \lamapi to build a new version of TURL-2K-red dataset which we called \textit{TURL-2K-red-LamAPI}.

\subsection{Training and implementation details}
\label{sec:training}

\textbf{Pre-training and model usage.}
For \turl we first replicated the pre-training with the same hyper\-parameters as specified by the authors in~\cite{Deng2022} but in a distributed setting on 4 80GB-A100 GPUs, following the findings in~\cite{goyal2018accurate}, then we fine-tuned it with the default hyper-parameters, matching the CEA results in the original paper. We use the open-source version of \tablellama made available on HuggingFace\footnote{https://huggingface.co/osunlp/TableLlama}.
\begin{figure*}[t!]
    \centering
    \includegraphics[width=\linewidth]{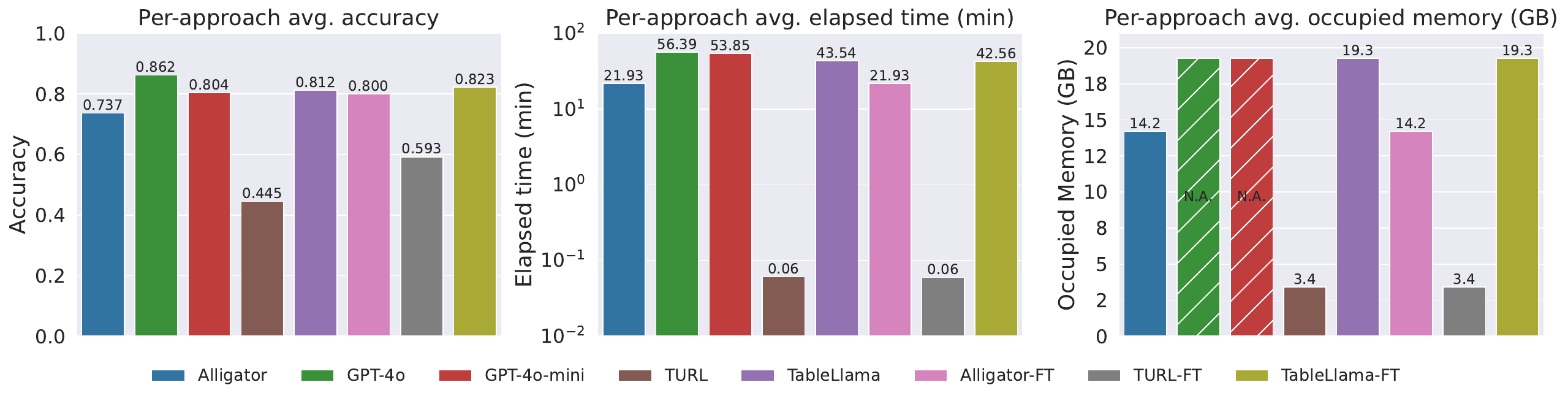}
    \caption{Per-approach average accuracy, elapsed time (min), and occupied memory (GB). The time reported is the elapsed time to link a mention, without considering the ER time. The occupied memory is the occupied GPU memory for \turl and \tablellama, while is the ``Virtual Memory Size'' for \alligator and \dagobah. \{Approach\}-FT represents the respective approach fine-tuned in MOOD setting (Sec.~\ref{sec:distribution-aware}). ``O.O.M.'' stands for Out-Of-Memory.}
    \label{fig:avgs-per-approach}
\end{figure*}
\begin{table*}[t!]
\scriptsize
\caption{Performance of the four algorithms on our test data, denoted as ``Accuracy\textsuperscript{\scalebox{.7}{Data Typology}}'', where the superscripts represent the data typology:
\textsuperscript{\scalebox{.7}{IN}}~\textit{in-domain},
\textsuperscript{\scalebox{.7}{MOOD}}~\textit{moderately out-of-domain},
\textsuperscript{\scalebox{.7}{OOD}}~\textit{out-of-domain}, and
\textsuperscript{\scalebox{.7}{MOOD$\rightarrow$IN}} indicates the Data Typology transition from MOOD to IN after the model has been fine-tuned in MOOD setting. The best model is highlighted in \textbf{bold}, while the second-best model is \underline{underlined}.
\{Dataset\}-red means that the specific dataset has been reduced as explained in Section~\ref{sec:datasets}. ``O.O.M.'' stands for Out-Of-Memory.}
\resizebox{\textwidth}{!}{
\begin{tabular}{llllllllll}
\toprule
\textbf{Dataset} & \textbf{\turl} & \textbf{\tablellama} & \textbf{\alligator} & \textbf{\dagobah} & \textbf{GPT-4o} & \textbf{GPT-4o-mini} & \textbf{\turl-FT} & \textbf{\tablellama-FT} & \textbf{\alligator-FT} \\
\midrule
SemTab2022 2T-red &
0.1343\textsuperscript{\scalebox{.7}{MOOD}} &
0.8243\textsuperscript{\scalebox{.7}{MOOD}} &
0.7156\textsuperscript{\scalebox{.7}{IN}} &
N.A.\textsuperscript{\scalebox{.7}{IN}} &
\textbf{0.8783}\textsuperscript{\scalebox{.7}{UNK}} &
0.7907\textsuperscript{\scalebox{.7}{UNK}} &
0.3323\textsuperscript{\scalebox{.7}{MOOD$\rightarrow$IN}} &
0.8400\textsuperscript{\scalebox{.7}{MOOD$\rightarrow$IN}} &
\underline{0.8531}\textsuperscript{\scalebox{.7}{IN}} \\
SemTab2022 HT-R1 &
0.3997\textsuperscript{\scalebox{.7}{MOOD}} &
0.7873\textsuperscript{\scalebox{.7}{MOOD}} &
\textbf{0.8890}\textsuperscript{\scalebox{.7}{IN}} &
0.7413\textsuperscript{\scalebox{.7}{IN}} &
\underline{0.8542}\textsuperscript{\scalebox{.7}{UNK}} &
0.8087\textsuperscript{\scalebox{.7}{UNK}} &
0.7454\textsuperscript{\scalebox{.7}{MOOD$\rightarrow$IN}} &
0.8001\textsuperscript{\scalebox{.7}{MOOD$\rightarrow$IN}} &
0.8165\textsuperscript{\scalebox{.7}{IN}} \\
SemTab2022 HT-R2 &
0.2763\textsuperscript{\scalebox{.7}{MOOD}} &
0.6619\textsuperscript{\scalebox{.7}{MOOD}} &
\textbf{0.8218}\textsuperscript{\scalebox{.7}{IN}} &
0.6289\textsuperscript{\scalebox{.7}{IN}} &
\underline{0.7516}\textsuperscript{\scalebox{.7}{UNK}} &
0.6565\textsuperscript{\scalebox{.7}{UNK}} &
0.6018\textsuperscript{\scalebox{.7}{MOOD$\rightarrow$IN}} &
0.6778\textsuperscript{\scalebox{.7}{MOOD$\rightarrow$IN}} &
0.7212\textsuperscript{\scalebox{.7}{IN}} \\
SemTab2023 WikidataTables-R1 &
0.3391\textsuperscript{\scalebox{.7}{MOOD}} &
0.7426\textsuperscript{\scalebox{.7}{MOOD}} &
\textbf{0.8248}\textsuperscript{\scalebox{.7}{IN}} &
0.7279\textsuperscript{\scalebox{.7}{IN}} &
0.8034\textsuperscript{\scalebox{.7}{UNK}} &
0.7554\textsuperscript{\scalebox{.7}{UNK}} &
0.7061\textsuperscript{\scalebox{.7}{MOOD$\rightarrow$IN}} &
0.7530\textsuperscript{\scalebox{.7}{MOOD$\rightarrow$IN}} &
\underline{0.8086}\textsuperscript{\scalebox{.7}{IN}} \\
\midrule
SemTab2021 BioDiv-red &
0.8109\textsuperscript{\scalebox{.7}{OOD}} &
0.9513\textsuperscript{\scalebox{.7}{OOD}} &
0.5674\textsuperscript{\scalebox{.7}{OOD}} &
N.A.\textsuperscript{\scalebox{.7}{OOD}} &
\textbf{0.9748}\textsuperscript{\scalebox{.7}{UNK}} &
0.9497\textsuperscript{\scalebox{.7}{UNK}} &
0.6347\textsuperscript{\scalebox{.7}{OOD}} &
\underline{0.9610}\textsuperscript{\scalebox{.7}{OOD}} &
0.8547\textsuperscript{\scalebox{.7}{OOD}} \\
\midrule
TURL-2K-red-LamAPI &
0.7118\textsuperscript{\scalebox{.7}{IN}} &
\underline{0.9051}\textsuperscript{\scalebox{.7}{IN}} &
0.6017\textsuperscript{\scalebox{.7}{MOOD}} &
O.O.M\textsuperscript{\scalebox{.7}{MOOD}} &
\textbf{0.9095}\textsuperscript{\scalebox{.7}{UNK}} &
0.8629\textsuperscript{\scalebox{.7}{UNK}} &
0.5347\textsuperscript{\scalebox{.7}{IN}} &
0.9045\textsuperscript{\scalebox{.7}{IN}} &
0.7456\textsuperscript{\scalebox{.7}{MOOD$\rightarrow$IN}} \\
\bottomrule
\end{tabular}%
}
\label{tab:performances}
\end{table*}
\alligator is pre-trained on different SemTab datasets before 2022 as in the original paper\footnote{As reported in Table~\ref{tab:data-pretrain}, the 2T dataset is one of the datasets used to pre-train \alligator. Even though we have 2T in our test data, those two datasets come from different rounds and years of the SemTab challenge.}. We refer to Table~\ref{tab:data-pretrain} for details about the datasets used for pre-train. \dagobah is run with the default hyper\-parameters as specified in the corresponding GitHub repository\footnote{https://github.com/Orange-OpenSource/Table-Annotation}.

\textbf{MOOD Fine-tuning}. We remind that we consider two evaluation settings, given the same test datasets (cfr. Table~\ref{tab:data-stats}): 1) the approaches are tested based on their pre-trained state; 2) \turl, \tablellama and \alligator are first fine-tuned on their MOOD data (cfr. Table~\ref{tab:data-stats} and Table~\ref{tab:data-wrt-data-type}) and subsequently tested. For the fine-tuning of \turl the default hyperparameters have been adopted as specified by the authors in~\cite{Deng2022}. For \tablellama we have employed the findings in~\cite{ibrahim2024simple}, \ie re\-warming the learning rate from $\eta_0=0.0$ to $\eta_{max}=2\text{e-}5$ for $0.5\%$ of the training iterations, then re\-decaying it with a cosine scheduler to reach $\eta_{min}=0.1\cdot\eta_{max}$ at the end of 2-epochs training, stopping it after 1 epoch due to clear signs of overfitting. Due to limited resources and budget \tablellama has been fine-tuned with LoRA~\cite{hu2022lora} following~\cite{Chen2024} with a micro batch-size$=1$, $64$ gradients accu\-mulation steps, LoRA\-rank $=8$, LoRA-$\alpha=16$, without any dropout or weight-decay. Finally, we have fine-tuned the \alligator's second model (RN) with the default hyperparameters as specified in~\cite{Avogadro2023}.

\textbf{Evaluation and inference.} Following our primary objective, all selected approaches were evaluated on the same test data (refer to Table~\ref{tab:data-stats}). Both \tablellama and \turl were run with a batch size of 1, though this setting has distinct interpretations: for \turl, it means processing one table at a time, while for \tablellama, it means processing a single mention at a time. The GPT-4o and GPT-4o-mini models were also run with \tablellama-style prompts, meaning they process one mention at a time. In contrast, \alligator was run with its default configuration, using 8 parallel processes, each handling up to 50 rows per table. Lastly, \dagobah was executed with its default hyperparameters, as specified in its \href{https://github.com/Orange-OpenSource/Table-Annotation/tree/main}{official GitHub repository}.   

\textbf{Technical infrastructure.} Both test and fine-tuning for \tablellama and \turl has run on the Azure \href{https://learn.microsoft.com/en-us/azure/virtual-machines/sizes/gpu-accelerated/nca100v4-series?tabs=sizebasic}{NC24ads\_a100\_v4}; \alligator on an Intel(R) Xeon(R) Gold 6238R CPU @ 2.20GHz with $64$ cores and $96$GB of RAM, while \dagobah on an Intel(R) Xeon(R) CPU E5-2650 0 @ 2.00GHz with $32$ cores and $96$GB of RAM.

\section{Results and discussion}
\label{sec:evaluation}
We first focus on the main results, then we discuss evidence from ablation studies.

\subsection{Main results}
\label{subsec:main-res}
Figure~\ref{fig:avgs-per-approach} reports the per-approach average accuracy, with the average computed over the test datasets (cfr. Table~\ref{tab:data-stats}), while Table~\ref{tab:performances} reports the detailed performances achieved by the four different approaches on the test data. We observe that \alligator excels on HT-R1, HT-R2 and WikidataTablesR1 SemTab datasets, it performs poorly on both BioDiv-red and TURL-2K-red-LamAPI while performing discretely on 2T-red dataset w.r.t. \tablellama. The opposite is observed for \tablellama and \turl, with \tablellama achieving generally higher ac\-curacy than \turl. This trend can be explained by the fact that \turl, \tablellama and \alligator encode different inductive biases: features computed by \alligator aim to capture signals coming from both the underline \ac{kg} and the syntactical relatedness between a mention and its candidates; GTUM models like \turl and \tablellama on the other hand capture more semantic and high-level features, lacking the direct interaction with the \ac{kg}. Moreover, the datasets they excel on are the ones considered \textit{in-domain} (IN). Surprisingly, both \turl and \tablellama excel on BioDiv, even though it's considered OOD for both approach\-es: we hypothe\-sise that the enormous and general pre-training knowledge these models retain explains this excellence. The poor performance achieved by \turl on all the SemTab data, conside\-red MOOD for it, can be explained by the fact that \turl is already fine-tuned on web tables, which are generally coming from a different data distribution than the ones of the SemTab challenge. Intere\-stingly, we observed that both \tablellama and \turl under\-performs them\-selves on the TURL-2K-LamAPI dataset: we argue that the drop in perfor\-mances, especially for \turl, is due to the increased number of candidates we have retrieved with LamAPI (cfr. the ablation study in section~\ref{subsec:ablations}). Regarding the GPT family, GPT-4o is, on average, the best-performing model, with GPT-4o-mini that is on par with the MOOD fine-tuned version of \alligator.

\begin{figure}[t!]
  \Description{}
  \includegraphics[width=\linewidth]{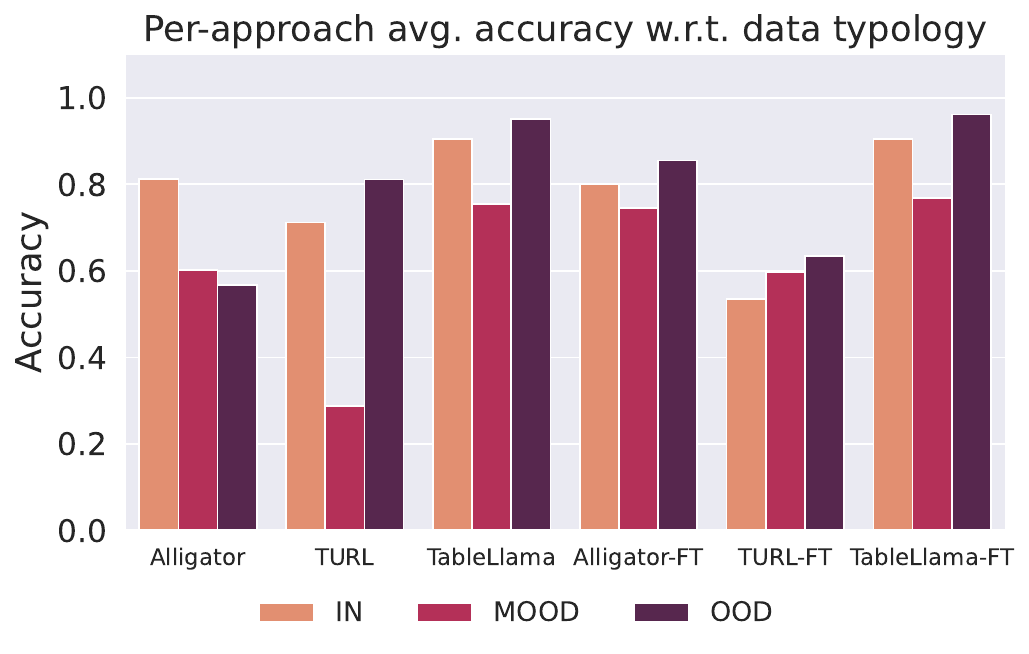}
  \caption{Per-approach accuracy with respect to data typology, as defined in Section~\ref{sec:distribution-aware}. The average is computed over datasets sharing the same data typology.}
  \label{fig:accuracy-dt}
\end{figure}

\textbf{MOOD fine-tuning}. Table~\ref{tab:performances} reports also the performances of \turl, \tablellama and \alligator after MOOD fine-tuning. Thanks to the ``continual fine\-tuning" inspired by~\cite{ibrahim2024simple} \tablellama slightly increases its perfor\-mances on (previously) MOOD and OOD (BioDiv-red) data, with no clue of suffering from catastrophic forgetting on its IN data (TURL-2K-red-LamAPI). \turl gains the most from the MOOD fine-tuning ($\approx$$+36\%$ on Wikidata\-Tables-R1) but suffers a severe drop in performance on both IN and OOD data, confirming the catastrophic forgetting suffered by models after fine-tuning~\cite{goodfellow2015catastrophicforgetting}. \alligator achieves great performances on 2T-red, BioDiv-red and TURL-2K-red-LamAPI after MOOD fine-tuning, while suffering a contained drop in performance (apart from HT-R2, where it loses 10\%), demonstrating a better generalization pattern than \turl. The average performance with respect to the data typology is depicted in Figure~\ref{fig:accuracy-dt}, which further demonstrates these findings.

\textbf{Computational performance}. Figure~\ref{fig:avgs-per-approach} also reports the average elapsed time (center) for the six tested approaches and the average occupied memory (right) by each approach (\turl and \tablellama's occupied memory refers to the GPU memory, while the \alligator and \dagobah's memory refers to the overall ``Virtual Memory Size''). Note that time is on a logarithmic scale and is considered as the time employed by an approach to link a mention without considering the ER time\footnote{The GPT-4o and GPT-4o-mini time is constrained by the limits imposed by \href{https://platform.openai.com/docs/guides/rate-limits.}{OpenAI API rate limits}}. If \turl is the fastest and the lightest from both time and occupied memory perspectives and \tablellama is the slowest and heaviest, \alligator strikes as a good compromise w.r.t. to execution time and occupied memory. \dagobah, on the other hand, is completely out-of-bounds, especially if one is bounded by resource or budget constraints.

\subsection{Ablation studies}
\label{subsec:ablations}
\begin{figure}[t!]
    \Description{Accuracies achieved by \tablellama and \turl w.r.t. the number of candidates.}
    \includegraphics[width=\columnwidth]{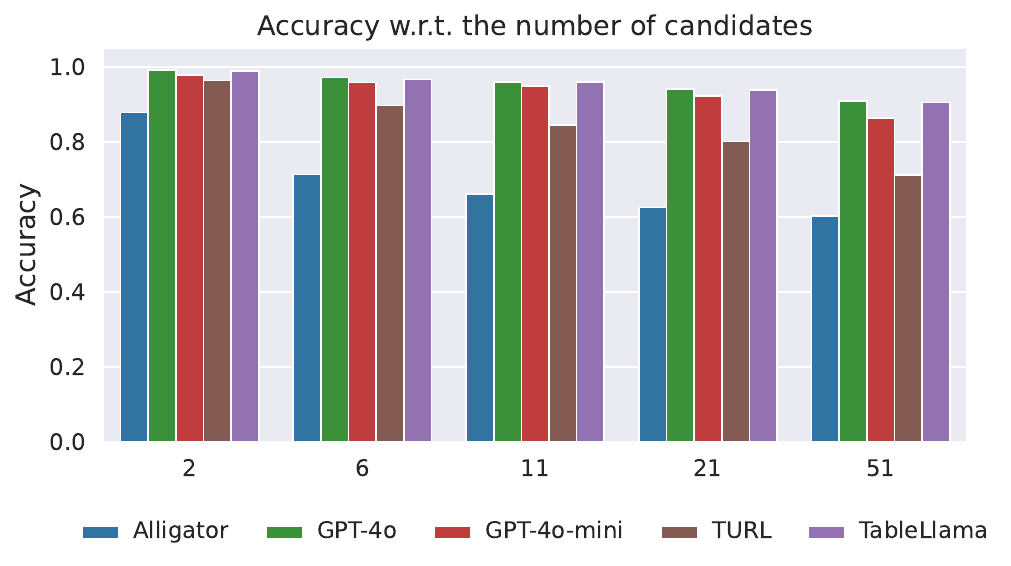}
    \caption{Accuracies achieved by \alligator, GPT4o, \tablellama and \turl w.r.t. the number of candidates on the TURL-2K-red-LamAPI dataset.}
    \label{fig:accuracy-candidates}
\end{figure}
\begin{figure}[t!]
  \Description{Accuracies achieved by \tablellama and \turl w.r.t. the presence of table's metadata.}
  \includegraphics[width=\linewidth]{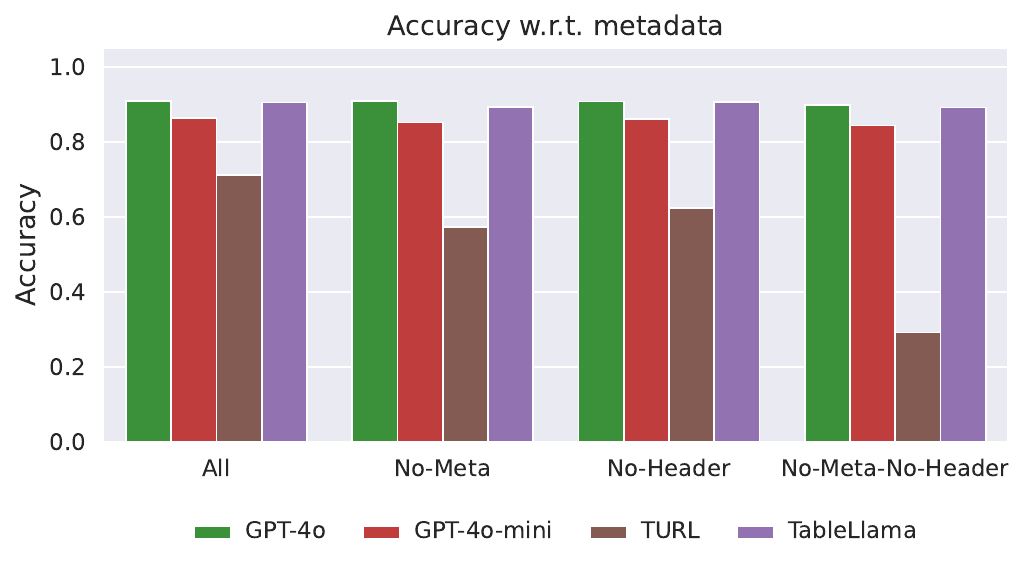}
  \caption{Accuracies achieved by GPT4o, \tablellama and \turl w.r.t. the presence of table's metadata on the TURL-2K-red-LamAPI dataset.}
  \label{fig:accuracy-metadata}
\end{figure}

To explain the drop in performances observed for \turl and \tablellama on the TURL-2K-red-LamAPI dataset, we have measured the accuracy w.r.t. the number of candidates per mention, with the intuition that the higher the number of candidates the lower the performances. Figure~\ref{fig:accuracy-candidates} reports the accuracy of \turl, \tablellama, GPT-4o and GPT-4o-mini given a different number of candidates per mention, also considering the correct one. Our intuitions are empirically confirmed by observing a drop in performance for both approaches, with a more severe one for \turl. This could happen because \turl aggregates the candidates of every mention in a table and passes them through a Transformer, increasing the context length to at most $O(NK)$, where $N$ is the number of mentions in a table and $K$ is the maximum number of candidates retrieved per mention. We ran an additional ablation study to measure the impact of the table's metadata (\eg the title of the Wikipedia page the table is found in, the section title, or the table caption to name a few) on the final accuracy achieved by \turl, \tablellama, GPT-4o, and GPT-4o-mini, testing on the TURL-2K-red-LamAPI dataset. We ran experiments with i) \textit{No-Meta}, \ie a setting where the page title, section title, and the table caption are removed; ii) \textit{No-Header}, \ie the table header is changed to $[\text{col}0, \text{col}1, ..., \text{col}N]$ and iii) \textit{No-Meta-No-Header}, \ie the combination of both i) and ii).
From Figure~\ref{fig:accuracy-metadata}, we observe that \turl is heavily dependent on the table metadata, with a severe drop in performance when both metadata and header are removed. On the other hand, \tablellama and GPT-4o are almost unaffected by removing metadata from the prompt, indicating a higher generalisation capability and a greater focus on the context provided by the table itself rather than by the metadata.

\subsection{Discussion}
\begin{figure}[t!]
  \Description{}
  \includegraphics[width=\linewidth]{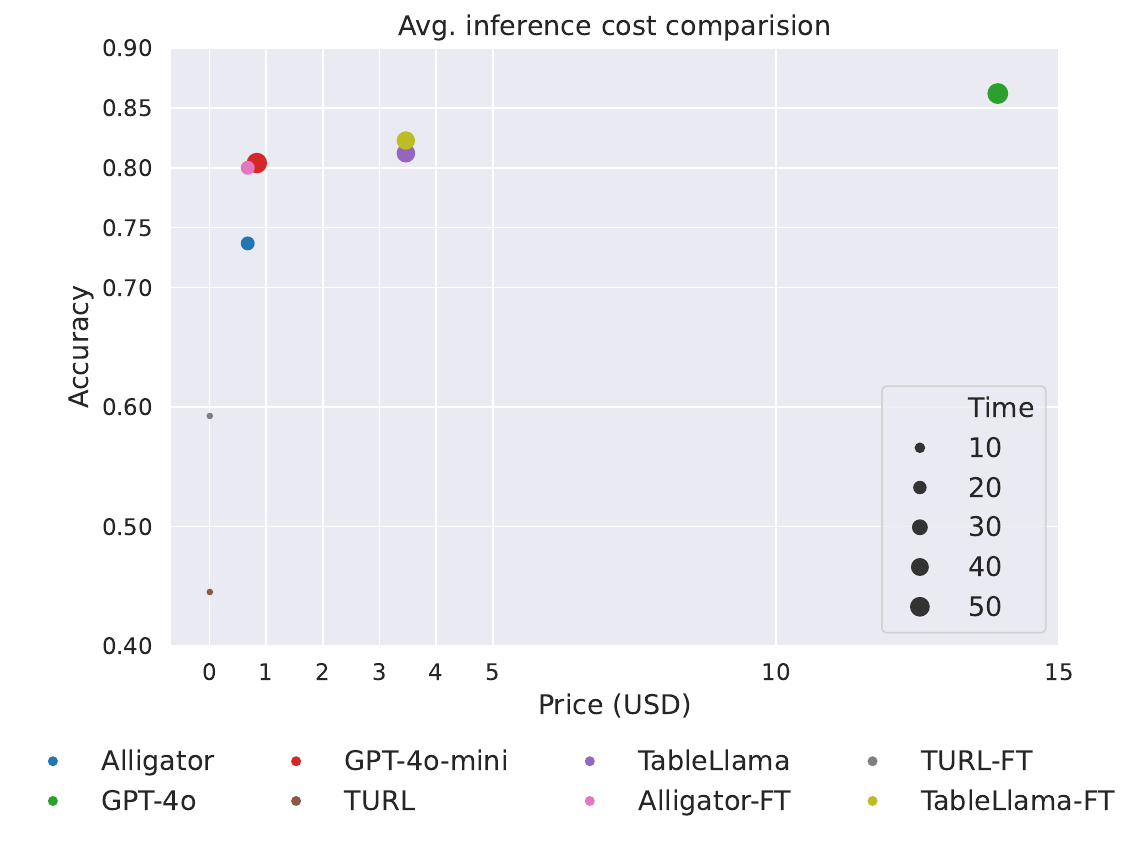}
  \caption{Per-approach accuracy over price, on average. The price is computed given the average time spent (in minutes) to run the evaluation over all the test data.}
  \label{fig:accuracy-price}
\end{figure}

Generative GTUM approaches with a high number of parameters have interesting properties in terms of accuracy, generalisa\-tion, and robustness to the number of candidates that are processed and the table metadata, especially on MOOD and OOD data, as found in Sections~\ref{subsec:main-res} and~\ref{subsec:ablations}; however, specific STI approaches trained on in-domain data, and optionally fine-tuned on moderately-out-of-domain data, using a tiny fraction of these parameters can still be on par with generalistic approaches while being faster and occupying less memory. 
% these results may suggest that generative GTUM approaches are at the moment the best choice for processing small tables, while more specific entity disambiguation approaches may still be a better fit for applications on large business data.
Generative approaches like GPT-4o and \tablellama are more promising than encoder-based ones in terms of performance, with a negligible risk of hallucinations; however, the comparison considered models of uncomparable size (7B vs 300M for \tablellama and \turl resp.), with TURL being the fastest approach among those that were tested. 
Regarding scalability, long context allows the encapsulation of large tables with a high number of candidates, but some tables cannot fit in the context, and when they do, the computation time and occupied memory increase proportionally to the table size: considering a small context table (few rows above and below the row containing the mention to be linked) to be fed to the model could be effective.
On the budget side, training models of the size of \tablellama has still enormous costs (48 A100-80GB for 9 training days), so devising generative methods based on smaller LLMs, \eg~Phi \cite{abdin2024phi3}, could be an interesting research direction, although more sophisticated approaches may be needed to achieve the same level of generalisation and reliability.
Moreover, Figure~\ref{fig:accuracy-price} shows the accuracy achieved by the tested approaches, both before and after MOOD fine-tuning, with respect to the average price\footnote{For \tablellama, \turl and \alligator, prices are estimated using the \href{https://azure.microsoft.com/en-us/pricing/calculator/}{Azure Pricing Calculator}, selecting the \href{https://learn.microsoft.com/en-us/azure/virtual-machines/sizes/gpu-accelerated/nca100v4-series?tabs=sizebasic}{NC24ads\_a100\_v4} for running GPU inference with \tablellama and \turl, and \href{https://learn.microsoft.com/en-us/azure/virtual-machines/sizes/general-purpose/dpldsv5-series?tabs=sizebasic}{D48plds\_v5} for running CPU inference and hosting \alligator. GPT-4o and GPT-4o-mini prices are estimated from the \href{https://openai.com/api/pricing/}{OpenAI prices page}, considering a token as being equal to \href{https://help.openai.com/en/articles/4936856-what-are-tokens-and-how-to-count-them}{4 characters}.} paid to host machines to run inference on all the test data. The figure shows how GPT-4o, while being the best-performing model on average, is also the most expensive one; GPT-4o-mini, \tablellama, \tablellama-FT, and \alligator-FT are comparable in terms of accuracy and price paid, although \tablellama costs is slightly higher. These results suggest that generative GTUM approaches are best suited when the tables to be annotated contain rich, highly semantic textual data. GPT-4o is the most accurate among these methods but also the most expensive. GPT-4o-mini offers a fair compromise between accuracy, cost, and computational demand. On the other hand, \alligator is more appropriate when signals from the knowledge graph (KG) have a significant impact on the final accuracy, especially when time and budget are limited. When fine-tuned, \alligator performs comparably to GPT-4o-mini and \tablellama in terms of accuracy with a reduced cost.
Finally, \tablellama and \turl are not NIL aware and \tablellama and GPT-4o, in particular, cannot return confidence scores associated with cell annotations, which may be useful when using these approaches to support revision: devising methods to estimate the uncertainty of labels computed by generative models may be an interesting research direction.

\section{Conclusions and Future Works}
\label{sec:conclusions}

In this study, we addressed the lack of comprehensive comparative analyses among different approaches to Semantic Table Interpretation (STI), focusing specifically on the Cell Entity Annotation (CEA) task. We selected four representative methods—including large generative Large Language Models (LLMs) and specialized STI models—and evaluated them under common conditions. Our evaluation spanned multiple dimensions: accuracy, generalisation, execution time, memory usage, and cost. We tested these methods across various scenarios, \ie ``\textit{in-domain}", ``\textit{out-of-domain}" and ``\textit{moderately out-of-domain}".
% \ac{llms} pre\-trained with a vast amount of data have been applied to \ac{sti} and \ac{cea}, complementing previous approaches based on heuristic and featured-based \ac{ml} approaches. However, these dif\-fe\-rent families of approaches have not been exhaustively examined on a common ground. In this work, we tackled this gap by selecting four represen\-tative approaches and comparatively evaluating them in terms of accuracy, generalisability, time, memory and price re\-qui\-re\-ments to better study their strengths and limitations, as well as their potential applications to different scenarios. We defined different evaluation settings, \ie ``\textit{in-domain}", ``\textit{out-of-domain}" and ``\textit{moderately out-of-domain}" for a better analysis of generalisability.
Our experiments suggest that approaches based on a large genera\-tive \ac{llm} excels in accuracy and ge\-ne\-ra\-li\-sa\-tion, as demonstrated by the results on MOOD and OOD data (cfr. Table~\ref{tab:performances} and Figure~\ref{fig:accuracy-dt}), at the price of an increased execution and training time. \turl, an encoder-only model based on TinyBERT, is the most efficient, but lacks on generalisation capabilities. Moreover, our experiments suggest that specific \ac{sti} models like \alligator, despite their limited number of parameters, can still be a viable option when resources and budget are limited, with performance improvements obtained after fine-tuning. Finally, GPT-4o, while being the most accurate, is also the most expensive, with GPT-4o-mini that represents a better trade-off with respect to accuracy and cost.  
Future works include the possibility to train a smaller and cheaper \ac{llm}-based table model, \eg Phi~\cite{abdin2024phi3}, while enabling also the handling of NIL entities and a score associated with cell annotations. Additionally, we aim to augment LLM-based approaches with advanced prompting techniques like In-Context Learning or Chain-of-Thought reasoning. This enhancement could leverage the inherent reasoning capabilities of LLMs and facilitate human-in-the-loop interactions to guide the model's behavior. Another potential direction is to modify models like \turl and \tablellama to operate at the cell level rather than the table level. This change could reduce memory consumption and allow for the application of standard data augmentation techniques, potentially improving both performance and efficiency.

\section{Acknowledgments}

This work has been partially funded by the European innovation action enRichMyData (HE 101070284) and the Italian PRIN project Discount Quality for Responsible Data Science: Human-in- the-Loop for Quality Data (202248FWFS) funded by the European Community - Next Generation EU.

\input{acronym.tex}

\bibliographystyle{splncs04}
\bibliography{biblio}

\end{document}

%% file: tables/pre_training_statistics.tex
\begingroup

\begin{table*}[ht!]
% \centering
\scriptsize
\caption{Statistics of the datasets used to pre-train the selected approaches}
\label{tab:data-pretrain}
    \resizebox{0.8\linewidth}{!}{%
    \begin{tabular}{lllll}
    
    \toprule
    % \rowcolor[HTML]{DDDDDD}
    \textbf{Approach} & \textbf{Dataset} & \textbf{Entities} & \textbf{Source} & \textbf{Entities Domain}\\
    \midrule
    % \rowcolor[HTML]{F3F3F3}
    \multirow{7}{*}{\textbf{\alligator}~\cite{Avogadro2023}} & SemTab2021 - R2 & 47.4K & Graph Queries & \multirow{7}{*}{Cross} \\
    % \rowcolor[HTML]{FFFFFF}
    & SemTab2021 - R3 & 58.9K & Graph Queries & \\
    % \rowcolor[HTML]{F3F3F3}
    & SemTab2022 - R2 (2T) & 994.9K & Graph Queries / Web Tables & \\
    % \rowcolor[HTML]{FFFFFF}
    & SemTab2020 - R4 & 667.2K & Graph Queries & \\
    % \rowcolor[HTML]{F3F3F3}
    & SemTab2019 - R3 & 390.4K & Graph Queries & \\
    % \rowcolor[HTML]{FFFFFF}
    & SemTab2019 - R1 (T2D) & 8K & Graph Queries / Web Tables & \\
    % \rowcolor[HTML]{F3F3F3}
    & Total & 2.16M & & \\
    \midrule
    % \rowcolor[HTML]{FFFFFF}
    \textbf{\turl}~\cite{Deng2022} & TURL WikiTable w/o WikiGS~\cite{efthymiou2017wikigs} & 1.23M & Web Tables & Cross \\
    \midrule
    % \rowcolor[HTML]{F3F3F3}
    \textbf{\tablellama}~\cite{Zhang2023} & TableInstruct~\cite{Zhang2023} & 2.6M & Web Tables & Cross \\
    \bottomrule
    \end{tabular}%
    }
\end{table*}

\endgroup

%% file: tables/datasets_statistics.tex
\begingroup

\begin{table*}[ht!]
  \caption{Statistics of the datasets used to fine-tune and evaluate models. \textdagger indicates datasets that have been sub-sampled so that they can be entirely processed by TableLlama within its 8192 context. \{Dataset\}-red means that the specific dataset has been reduced as explained in Section \ref{sec:datasets}.}
    \label{tab:data-stats}
    \resizebox{0.8\linewidth}{!}{%
    \begin{tabular}{llrccr}
      \toprule
      
      {\textbf{Dataset}} & {\textbf{Split}} & {\textbf{Tables}} & \shortstack{\textbf{Cols (min $|$ max $|$ $\Bar{x}$)}} & \multicolumn{1}{l}{\textbf{Rows (min $|$ max $|$ $\Bar{x}$)}} &  {\textbf{Entities}} \\

      \midrule

      & Train\textdagger & \numprint{91} &\numprint{1} $|$ \numprint{8} $|$ \numprint{4.86} & \numprint{5} $|$ \numprint{369} $|$ \numprint{98.61} & \numprint{14674} \\

      \multirow{-2}{*}{SemTab2022 2T-red} & Test\textdagger & \numprint{26} & \numprint{1} $|$ \numprint{8} $|$ \numprint{4.65} & \numprint{7} $|$ \numprint{264} $|$ \numprint{74.58} & \numprint{4691} \\

      & Train & \numprint{3691} & \numprint{2} $|$ \numprint{5} $|$ \numprint{2.56} & \numprint{4} $|$ \numprint{8} $|$ \numprint{5.68} & \numprint{26189} \\
      
      \multirow{-2}{*}{SemTab2022 HT-R1} & Test & \numprint{200} & \numprint{2} $|$ \numprint{5} $|$ \numprint{2.59} & \numprint{4} $|$ \numprint{8} $|$ \numprint{5.74} & \numprint{1406} \\

      & Train\textdagger & \numprint{4344} & \numprint{2} $|$ \numprint{5} $|$ \numprint{2.56} & \numprint{4} $|$ \numprint{8} $|$ \numprint{5.57} & \numprint{20407} \\
      
      \multirow{-2}{*}{SemTab2022 HT-R2} & Test & \numprint{426} & \numprint{2} $|$ \numprint{5} $|$ \numprint{2.53} & \numprint{4} $|$ \numprint{8} $|$ \numprint{5.56} & \numprint{1829} \\

       & Train & \numprint{9917} & \numprint{2} $|$ \numprint{4} $|$ \numprint{2.51} & \numprint{3} $|$ \numprint{11} $|$ \numprint{5.65} & \numprint{64542} \\

      \multirow{-2}{*}{SemTab2023 WikidataTables-R1} & Test & \numprint{500} & \numprint{2} $|$ \numprint{4} $|$ \numprint{2.46} & \numprint{3} $|$ \numprint{11} $|$ \numprint{6.95} & \numprint{4247} \\

      \midrule

      SemTab2021 BioDiv-red & Test\textdagger & \numprint{11} & \numprint{1} $|$ \numprint{26} $|$ \numprint{17.45}
      &\numprint{26} $|$ \numprint{100} $|$ \numprint{58.90} & \numprint{1232} \\

      \midrule
      
       TURL-120k & Train & \numprint{13061} & \numprint{1} $|$ \numprint{43} $|$ \numprint{5.43} & \numprint{1} $|$ \numprint{624} $|$ \numprint{12.98} & \numprint{120000} \\

      TURL-2k-red & Test & \numprint{1295} & \numprint{1} $|$ \numprint{14} $|$ \numprint{1.03} & \numprint{6} $|$ \numprint{257} $|$ \numprint{32.95} & \numprint{1801} \\
      \bottomrule
    \end{tabular}
    }
\end{table*}

\endgroup

%% file: tables/datasets_domain.tex
\begingroup
% \dashlinedash=1pt
% \dashlinegap=1pt
% \setlength{\tabcolsep}{5pt}
% \renewcommand{\arraystretch}{1.5}

\begin{table*}[ht!]
\scriptsize
\caption{Test datasets denomination, as specified in Section~\ref{sec:distribution-aware}. For every dataset, we report its source, the domain of the entities contained, and the characterization as In-Domain (IN), Out-Of-Domain (OOD), and Moderately-Out-Of-Domain (MOOD) for each of the pre-trained models (see Table~\ref{tab:data-pretrain} for the pre-training datasets).
% also considering the source and the entity domain of the pre-trained models as specified in Table \ref{tab:dataset-charateristics}.
\{Dataset\}-red means that the specific dataset has been reduced as explained in Section \ref{sec:datasets}.}
\label{tab:data-wrt-data-type}
    \resizebox{0.8\linewidth}{!}{
    \begin{tabular}{lllllll}
    
    \toprule
    
     &  &  & \multicolumn{3}{c}{\textbf{Approaches}} \\
    
    {\multirow{-2}{*}{\textbf{Dataset}}} & {\multirow{-2}{*}{\textbf{Source}}} & {\multirow{-2}{*}{\textbf{Entities Domain}}} & \multicolumn{1}{l}{\textbf{Alligator}} & \multicolumn{1}{l}{\textbf{TURL}} & \multicolumn{1}{l}{\textbf{TableLlama}} \\
    
    \midrule
    
    {SemTab2022 2T-red} & {Graph Queries / Web Tables} & {Cross} & IN & MOOD & MOOD \\
    
    {SemTab2022 HT-R1} & {Graph Queries} & {Cross}  & IN & MOOD & MOOD \\
    
    {SemTab2022 HT-R2} & {Graph Queries} & {Cross}  & IN & MOOD & MOOD \\
    
    {SemTab2023 WikidataTables-R1} & {Graph Queries} & {Cross}  & IN & MOOD & MOOD \\
    \midrule
    
    {SemTab2021 BioDiv-red} & {Biodiversity Tables} & {Specific} & OOD  & OOD  & OOD  \\
    \midrule
    
    {TURL-120k} & {Web Tables} & {Cross}  & MOOD & IN & IN \\
    
    {TURL-2K-red} & {Web Tables} & {Cross}  & MOOD & IN & IN \\
    \bottomrule
    
    \end{tabular}%
    }
\end{table*}
\endgroup

%% file: acronym.tex
\begin{acronym}
\acro{ai}[AI]{Artificial Intelligence}
\acro{bow}[BOW]{Bag-Of-Words}
\acro{bkg}[BKG]{Background Knowledge Graph}
\acro{cta}[CTA]{Column-Type Annotation}
\acro{cea}[CEA]{Cell-Entity Annotation}
\acro{cpa}[CPA]{Columns-Property Annotation}
\acro{crf}[CRF]{Conditional Random Field}
\acro{dl}[DL]{Deep Learning}
\acro{el}[EL]{Entity Linking}
\acro{ed}[ED]{Entity Disambiguation}
\acro{er}[ER]{Entity Retrieval}
\acro{gui}[GUI]{Graphical User Interface}
\acro{gdbt}[GDBT]{Gradient Boosted Decision Tree classification model}
\acro{gs}[GS]{Gold Standard}
\acro{gss}[GSs]{Gold Standards}
\acro{gtum}[GTUM]{Generalistic Table Understanding and Manipulation}
\acro{kg}[KG]{Knowledge Graph}
\acro{kgs}[KGs]{Knowledge Graphs}
\acro{kb}[KB]{Knowledge Base}
\acro{kbs}[KBs]{Knowledge Bases}
\acro{lod}[LOD]{Linked Open Data}
\acro{cnea}[CNEA]{Cell-New Entity Annotation}
\acro{nlp}[NLP]{Natural Language Processing}
\acro{regex}[Regex]{Regular Expressions}
\acro{rdf}[RDF]{Resource Description Framework}
\acro{rml}[RML]{RDF Mapping Language}
\acro{rdfs}[RDFS]{Resource Description Framework Schema}
\acro{sota}[SOTA]{state-of-the-art}
\acro{sti}[STI]{Semantic Table Interpretation}
\acro{svm}[SVM]{Support Vector Machine}
\acro{ui}[UI]{User Interface}
\acro{ir}[IR]{Information Retrieval}
\acro{ml}[ML]{Machine Learning}
\acro{ne}[NE]{Named Entity}
\acro{ner}[NER]{Named Entity Recognition}
\acro{nel}[NEL]{Named Entity Linking}
\acro{nil}[NIL]{Not In Lexicon}
\acro{nn}[NN]{Neural Network}
\acro{lit}[LIT]{Literal}
\acro{crf}[CRF]{Conditional Random Field}
\acro{cnn}[CNN]{Convolutional Neural Network}
\acro{rnn}[RNN]{Recurrent Neural Network}
\acro{hnn}[HNN]{Hybrid Neural Network}
\acro{llm}[LLM]{Large Language Model}
\acro{llms}[LLMs]{Large Language Models}
\acro{lms}[LMs]{Language Models}
\acro{lm}[LM]{Language Model}
\acro{hdt}[HDT]{Hybrid Decision Tree}
\acro{pfsms}[PFSMs]{Probabilistic Finite-State Machines}
\acro{pgm}[PGM]{Probabilistic Graphical Model}
\acro{gft}[GFT]{Google Fusion Tables}
\acro{pois}[PoI]{Points of Interest}
\end{acronym}